\documentclass[draft]{agujournal2019}
\usepackage{url} %this package should fix any errors with URLs in refs.
\draftfalse

\journalname{JGR: Machine Learning and Computation}

\usepackage{amsmath} 
\usepackage{amssymb}
 \usepackage{soul}
\usepackage{graphicx} % Required for inserting images

\usepackage{enumitem}

\begin{document}

%%%%%%%%%

\title{Multidimensional Distributional Neural Network Output Demonstrated in Super-Resolution of Surface Wind Speed}

\authors{
    Harrison J. Goldwyn\affil{1},
    Mitchell Krock\affil{2},
    Johann Rudi\affil{3}, 
    Daniel Getter\affil{4},
    Julie Bessac\affil{1, 3}, 
    }

\affiliation{1}{National Renewable Energy Laboratory}
\affiliation{2}{University of Missouri}
\affiliation{3}{Virginia Tech}
\affiliation{4}{University of Southern California}

\correspondingauthor{Harrison J. Goldwyn}{hgoldwyn@nrel.gov} 

%%%%%%%%%%%%%%%%%%%%%%%%%%%%%%%%%%%%%%%%%%%%%%%
% KEY POINTS
%%%%%%%%%%%%%%%%%%%%%%%%%%%%%%%%%%%%%%%%%%%%%%%
%  List up to three key points (at least one is required)
%  Key Points summarize the main points and conclusions of the article
%  Each must be 140 characters or fewer with no special characters or punctuation and must be complete sentences

% Example:
% \begin{keypoints}
% \item	List up to three key points (at least one is required)
% \item	Key Points summarize the main points and conclusions of the article
% \item	Each must be 140 characters or fewer with no special characters or punctuation and must be complete sentences
% \end{keypoints}

\begin{keypoints}
\item Neural networks can be trained on multidimensional probabilistic loss functions to capture spatially correlated heteroscedastic uncertainty.
\item Closed-form distributional predictions enable fast sampling and interpretation of uncertainty in spatial neural network outputs.
\item Demonstrated framework enables closed-form uncertainty aware downscaling with potential extensions to complex distributional forms.
\end{keypoints}

%%%%%%%%%%%%%%%%%%%%%%%%%%%%%%%%%%%%%%%%%%%%%%%
%
%  ABSTRACT and PLAIN LANGUAGE SUMMARY
%
% A good Abstract will begin with a short description of the problem
% being addressed, briefly describe the new data or analyses, then
% briefly states the main conclusion(s) and how they are supported and
% uncertainties.

% The Plain Language Summary should be written for a broad audience,
% including journalists and the science-interested public, that will not have 
% a background in your field.
%
% A Plain Language Summary is required in GRL, JGR: Planets, JGR: Biogeosciences,
% JGR: Oceans, G-Cubed, Reviews of Geophysics, and JAMES.
% see http://sharingscience.agu.org/creating-plain-language-summary/)
%
%%%%%%%%%%%%%%%%%%%%%%%%%%%%%%%%%%%%%%%%%%%%%%%

\begin{abstract}
Accurate quantification of uncertainty in neural network predictions remains a central challenge for scientific applications involving high-dimensional, correlated data. While existing methods capture either aleatoric or epistemic uncertainty, few offer closed-form, multidimensional distributions that preserve spatial correlation while remaining computationally tractable. In this work, we present a framework for training neural networks with a multidimensional Gaussian loss, generating closed-form predictive distributions over outputs with non-identically distributed and heteroscedastic structure. Our approach captures aleatoric uncertainty by iteratively estimating the means and covariance matrices, and is demonstrated on a super-resolution example. We leverage a Fourier representation of the covariance matrix to stabilize network training and preserve spatial correlation. We introduce a novel regularization strategy---referred to as information sharing---that interpolates between image-specific and global covariance estimates, enabling convergence of the super-resolution downscaling network trained on image-specific distributional loss functions. This framework allows for efficient sampling, explicit correlation modeling, and extensions to more complex distribution families all without disrupting prediction performance. We demonstrate the method on a surface wind speed downscaling task and discuss its broader applicability to uncertainty-aware prediction in scientific models. 
\end{abstract}

\section*{Plain Language Summary}
Artificial neural networks (often called ``neural networks") are mathematical models inspired by networks of biological neurons that are simple to implement in software and tend to generate fast and accurate predictions in various settings. 
The downside of neural networks is that they output predictions without reporting their confidence, which can be risky in real-world applications.
In this study, we develop a framework for training neural networks: the network predicts a best guess along with its uncertainty around that guess in a way that accounts for the relationship between data at different points in space. 
We applied our method to the task of improving the resolution of surface wind speed simulations: turning cheap-to-generate coarse images into detailed output while reporting how reliable each part of the estimated high-resolution image is. 
Our method accounts for how the data varies across space and provides a fast way to generate samples with consistent data patterns learned by the network.
These results demonstrate a promising way to train neural networks to report their uncertainty in problems involving complex spatial data. 
This approach and its proposed extensions could help researchers and policy makers make better-informed decisions using fast and transparent tools. 
% \julie{cool!}
%%%%%%%%%%%%%%%%%%%%%%%%%%%%%%%%%%%%%%%%%%%%%%%%%%%%%%%%%%%%%%%%%%%%%
% MAIN BODY OF PAPER
%%%%%%%%%%%%%%%%%%%%%%%%%%%%%%%%%%%%%%%%%%%%%%%%%%%%%%%%%%%%%%%%%%%%%  

%%%%%%%%%%%%%%%%%%%%%%%%%%%%%%%%%%%%%%%%%%%%%%%%%%%%%%%%%%%%%%%%%%%%%
%% Intro (see \cite{Merizzi2024} for a good intro...)
%%%%%%%%%%%%%%%%%%%%%%%%%%%%%%%%%%%%%%%%%%%%%%%%%%%%%%%%%%%%%%%%%%%%%  
\section{Introduction}

Artificial neural networks have achieved widespread success across scientific disciplines, achieving state-of-the-art performance in tasks such as image and signal processing \cite{hu2018handbook}, molecular modeling \cite{schneider1998artificial}, fluid dynamics \cite{kutz2017deep}, and earth system science \cite{liu2016application}. 
Their capacity to learn complex nonlinear relationships from large datasets has enabled breakthroughs in both predictive accuracy and the discovery of underlying physical phenomena. 
As a result, neural networks (NN) are increasingly adopted not only as black-box predictors but also as flexible tools for augmenting or accelerating traditional scientific models \cite{faroughi2022physics}.

Despite the success of NNs, quantifying uncertainty is difficult. 
In high-stakes applications such as medical diagnosis, autonomous vehicles, and weather prediction, decisions based on point estimates alone can be misleading or unsafe because uncertainties rooted in both the data and modeling are not accounted for. 
The difficulty lies in the supervised training of deep neural networks, which is a complex optimization problem. 
Many training samples are needed to computationally learn a deterministic, nonlinear, and nonconvex mapping from input vectors to output vectors across the training dataset. 
The predictions output by a trained network are usually interpreted as point estimates. These single-point outputs do not describe any statistical properties across samples of the dataset despite such information being present in the training set. 

Designs for neural networks that generate probabilistic output have a long history \cite{specht1990probabilistic, cheng1994neural, mohebali2020probabilistic}, with early attempts using non-parametric estimation of the distributions costing high memory usage and slow inference time.
Many methods have since been explored, each with potential benefits and weaknesses. 
Most publications on uncertainty quantification (UQ) in deep neural networks can be classified as deterministic or stochastic methods for distributional inference \cite{gawlikowski2023survey}. 
Deterministic methods cover any approach where uncertainty in predicted parameters is estimated based on a single forward pass of a deterministic network. 
These methods have often relied on choosing distributional loss functions that ensure distributional parameters can be estimated.
Using a single network ensures computational efficiency, but comes at the cost of sensitivity to the network architecture, training procedure, and training data.
Stochastic methods include all implementations of neural networks where outputs are not uniquely determined by inputs. 
In this case, model parameters are treated explicitly as random variables. 
Stochastic methods often gain accuracy in distributional estimates at the cost of the increased computational complexity of optimizing stochastic functions. 

UQ methods from either of these classes may capture one of two primary types of uncertainty: aleatoric and epistemic. 
The term aleatoric is derived from the Latin description of \emph{something that depends on the throw of a dice}, and refers to any randomness inherent to a process that cannot be reduced with further data collection. Epistemic uncertainty is named from the Greek word for knowledge and refers to uncertainty rooted in modeling choices. Theoretically, epistemic uncertainty can be completely resolved if enough replicate data are available to determine the perfect model. 
Deterministic methods tend to adapt well to aleatoric uncertainty by attempting to specify the data-generating process. 
For stochastic methods, Bayesian neural networks are the archetypal quantification of epistemic uncertainty, with the fundamental update to traditional neural networks being the specification of a prior distribution on network weights and the propagation of those weight distributions through the network. 

It is the intent of this work to tackle the problem between these two classifications---treating distributional parameters for nontrivial, multidimensional, correlated distributions as random variables to be estimated by a neural network trained on a probabilistic loss function. 
As we detail below, our approach extends prior deterministic methods for NN UQ based on a single forward pass NN, to iterative learning of multidimensional distribution parameters. 
This framework relies on reinterpretation of the standard neural network loss function as a Bayesian posterior, which allows for closed-form inference of aleatoric uncertainty.
We leave epistemic uncertainty aside by choosing to estimate posterior parameters for spatial data that are conditional yet independently Gaussian distributed. 
This modeling choice serves as a step towards more complex multidimensional distributions in future work. 
This work presents a proof-of-concept framework for training a NN to predict parameters in a distributional loss function in the context of correlated, non-identically distributed spatial data. 
We refer to the model as capturing heteroscedastic (heterogeneity of variance) aleatoric uncertainty, because it predicts input-specific covariance parameters.  

This work was motivated by the need for efficient, uncertainty-aware downscaling methods in earth system modeling, where running high-resolution physics-based models provides accurate outputs but is often prohibitively expensive. 
Statistical downscaling techniques resolve this dilemma and have been adapted from the computer vision problems in super-resolution image enhancement. 
These methods have been implemented successfully, particularly those using deep learning architectures such as convolutional neural networks (CNNs) aided by residual learning \cite{dong2014learning, kim2016accurate, he2016deep, lim2017enhanced, ahn2018fast}, transformers \cite{liang2021swinir, conde2022swin2sr}, and a range of generative models including Diffusion models \cite{li2022srdiff}, and Generative Adversarial Networks along with their stochastic extensions \cite{ledig2017photo,wang2018esrgan,daust+m24}. 
Bayesian networks and generative models are capable of uncertainty quantification, but they tend to be computationally intensive, generally neglect spatial correlation, and may suffer from hallucinations \cite{oberdiek2023uqganunifiedmodeluncertainty,stengel2020adversarial,annau+cm23}.
We seek to retain the speed and scalability of traditional neural networks while also learning spatial statistics of reconstructed high-resolution fields.

This paper proposes a general framework for distributional output from neural networks by interpreting the loss function as a Bayesian posterior distribution, capturing heteroscedastic aleatoric uncertainty data that is sparsely sampled and non-identically distributed. 
We extend prior scalar approaches by training neural networks under a multidimensional Gaussian (MDG) loss on spatially correlated data, which yields a closed-form characterization of spatial uncertainty. 
Our framework is generally applicable to spatial prediction tasks as well as other NN application requiring uncertainty quantification over correlated multidimensional variables.  
We see the following advantages in our approach. First, fast sampling from predictive distributions facilitates statistical analysis with minimal computational cost. 
Second, the framework is interpretable because of an explicit representation of aleatoric correlation structure in Fourier space. %, offering transparency into learned spatial correlation lengths. 
Third, our approach offers better interpretability and significantly lower training complexity than generative models. 
We apply the framework to super-resolution of surface wind speed (SWS), which provides a representative case of spatially distributed data, where uncertainty quantification must cover local variance and spatial correlation. 
The MDG loss function successfully captures both the mean and spatial covariance of SWS. 
This approach extends previous scalar distributional models \cite{getter2024statistical}, which also were applied to SWS, to a correlated multidimensional setting, allowing the model to learn heteroscedastic aleatoric uncertainty across space. 

In what follows, we first introduce the framework in Section \ref{section:framework} and discuss the challenges that led us to our implemented approach. 
In Section \ref{section:cnn} we introduce the architecture of the CNN, specify the training conditions, and discuss the surface wind speed dataset used in our example application.  
In Section \ref{section:results}, we discuss our chosen performance metrics and step through results of all 3 stages of our proposed framework applied to super resolution of surface wind speed. 
We also present a statistical analysis of samples generated from predicted closed-form distributions, comparing results from an Image-Specific Covariance model to a Global Covariance model (homoscedastic limit). 
Section \ref{section:conclusion} reiterates the contributions of this work; and we close in Section \ref{section:future} with a discussion of directions for future work.

%%%%%%%%%%%%%%%%%%%%%%%%%%%%%%%%%%%%%%%%%%%%%%%%%%%%%%%%%%%%%%%%%%%%%
%% Approach
%%%%%%%%%%%%%%%%%%%%%%%%%%%%%%%%%%%%%%%%%%%%%%%%%%%%%%%%%%%%%%%%%%%%% 
\section{A Framework for Fitting NNs with multidimensional Distributional Loss Functions} \label{section:framework}

In this section, we introduce our framework for training a CNN to predict the mean and covariance for high-resolution output images assumed to be distributed according to conditional but independent MDG distributions.
Although the SWS data we examine here is timeseries data, we do no make use of temporal context \cite{9700735}. 
In this work, we develop a formalism for super-resolution of a single image. 
We will discuss the motivation and difficulties encountered in solving this problem in the context of super-resolved downscaling of SWS. Our proposed solution serves as a proof-of-concept for an iterative approach yielding complex closed-form distributional output from neural networks for heteroscedastic data.

\subsection{The Problem Statement}

The problem of training a neural network, defined by $  \mathbf{Y} = f(\mathbf{X}; \mathbf{W})$, is stated as finding weights $\mathbf{W}$ that minimize a loss function $L$ over inputs $\mathbf{X}$ and outputs $\mathbf{Y}$. 
The optimization,
\begin{equation}
    \mathbf{W}^* = \arg\min_{\mathbf{W}} \ {L}(f(\mathbf{X}; \mathbf{W}), \mathbf{Y})
\end{equation}
minimizes error in network output.
The loss is typically chosen as a low-dimensional computationally convenient scalar function, like the mean-squared error (MSE), that quantifies prediction error. 
Here we reinterpret the loss function as the Bayesian posterior distribution (or likelihood assuming a uniform prior). 
In what follows, we extend previous work taking this interpretation towards multidimensional and non-identically distributed data.

Given the task of predicting a probability distribution for spatially correlated data, we aim to estimate the 
mean $\boldsymbol{\mu} \in \mathbb{R}^k$ and the positive semi-definite covariance matrix $\mathbf{\Sigma} \in \mathbb{R}^{k,k}$ of the multidimensional Gaussian distribution,
\begin{equation}
\label{eq:mvn}
\mathrm{N}(\mathbf{Y}|\boldsymbol{\mu}, \mathbf{\Sigma}) = 
\sqrt{(2\pi)^{-k}|\mathbf{\Sigma}|^{-1}} \exp\left(
-\frac{1}{2} (\mathbf{Y} - \boldsymbol{\mu})^\mathrm{T}\mathbf{\Sigma}^{-1} (\mathbf{Y} - \boldsymbol{\mu})
\right).\end{equation}
To train a network on this distribution, we take the parameters 
% \hl{should mu and sigma be bold here:} 
$\{\boldsymbol{\mu}, \boldsymbol{\Sigma}\} = f(\mathbf{X}; \mathbf{W})$ to be functions of the network inputs and weights.
Then, we assume that solutions of interest maximize the process likelihood and the network weights are trained on the minimization of the negative logarithm of the multidimensional Gaussian, $\allowdisplaybreaks \displaystyle\mathbf{W}^* = \arg\min_{\mathbf{W}} \ L(\{\boldsymbol{\mu}_i, \boldsymbol{\Sigma}_i\}| \{\mathbf{Y}_i\})$. 
This loss function becomes, 
\begin{equation} \label{eq:ideal_loss}
    L(\{\boldsymbol{\mu}_i, \boldsymbol{\Sigma}_i\}| \{\mathbf{Y}_i\}) 
  = 
  \sum_{i \in \{\mathrm{img}\}}
  \left[
  \log(|\boldsymbol{\Sigma}_i|)
  +
  (\mathbf{Y}_i - \boldsymbol{\mu}_i)^T \boldsymbol{\Sigma}_i^{-1} (\mathbf{Y}_i - \boldsymbol{\mu}_i)
\right],
\end{equation}
where we assign a mean vector and covariance matrix to each image.
Solving this problem directly is not possible because we are effectively trying to estimate distributions from single samples. 

We may be tempted here to simplify the problem by returning to Eq.\ \eqref{eq:mvn} with the assumption of a global covariance: replacing $\boldsymbol{\Sigma}_i$ with
$\boldsymbol{\Sigma}_g$ for every image.
Unfortunately, the problem of covariance inversion remains difficult even in this homoscedastic limit. Inverting a matrix reaching the magnitude of 1000 $\times$ 1000 inversion is long-standing challenge intertwining numerical instability with expensive computational operations. 
Often linear algebra treatments or different modeling assumptions have to be made to overcome covariance inversion issues.

\subsection{Initial Implementation Challenges}

A significant preliminary finding that constrained our approach was the high sensitivity of the CNN training on the covariance matrix. 
This is a well-understood problem in the context of Gaussian Processes. 
\citeA{karvonen2023maximum} rigorously demonstrated that using maximum likelihood estimation to determine kernel hyperparameters can be ill-posed, meaning that even small perturbations in the data may cause large fluctuations in model predictions. 
This instability is compounded by the common occurrence of near-singular or ill-conditioned covariance matrices; such degeneracy can render covariance inversions numerically unreliable even when the computational expense of matrix inversion can be justified \cite{mohammadi2016analytic}. 
These issues require strong assumptions to be incorporated in the framework design and the use of regularization.

We implemented and tried training a CNN on the loss in Eq.\ \eqref{eq:ideal_loss} under numerous reduced representations of the covariance, including: parameterization by a spatially decaying correlation function, Cholesky decomposition to enforce invertibility, and empirical orthogonal basis function representation. 
None of these methods converged, either due to the degeneracy of the parameter space or the instability of the gradient descent through the covariance decomposition and inversion. 
These difficulties eventually lead us to an iterative approach, but our first key insights came from simplifying the problem down to training the network with fixed global covariance, 
\begin{equation} \label{eq:fixed_cov_loss}
    L(\{\boldsymbol{\mu}_i\} | \{\mathbf{Y}_i\}, \boldsymbol{\Sigma}_g\}) 
  = 
  \sum_{i \in \{\mathrm{img}\}}
  (\mathbf{Y}_i - \boldsymbol{\mu}_i)^T \boldsymbol{\Sigma}_g^{-1} (\mathbf{Y}_i - \boldsymbol{\mu}_i).
\end{equation}
Here, $\boldsymbol{\Sigma}_g$ was taken to be the sample covariance across the entire train and the test set $\boldsymbol{\Sigma}_g = \mathbf{Y}^T\mathbf{Y}/n_s$, where $\mathbf{Y}$ a matrix formed by vectorizing and stacking each image in the dataset as columns. 
This sample covariance was often singular, so to obtain a pseudoinverse, we truncated singular modes that had negligible contribution.

These results revealed two important findings. 
First, as singular mode number increased, the predicted high-resolution images tended to smooth out all spatial fine scales. 
In essence, the more accurate our model of the global sample covariance, the less detail the network could learn from low-resolution (LR) input images. 
We conjecture that this results from the images being non-identically distributed and enforcing a global covariance washed out the unique spatial structure of individual images. 
This,  we took as an encouragement to try estimating image-specific covariance matrices, and respect the dissimilarity between spatial structure image-to-image. 
Second, the results were extremely sensitive to the sample covariance, given that high-order singular modes has very little contribution to the overall variation in the data. 
We found this consistent with results presented by our ultimate approach detailed below. 

\subsection{An Iterative Framework for Image-specific Covariance Estimation}

To avoid the above issues, we implement two fundamental strategies:
First, we iteratively optimize for the grid-point means $\boldsymbol{\mu}_i$ and  for the image-specific covariance $\boldsymbol{\Sigma}_i$. 
But still, we face the issue of estimating distributional parameters from single samples if the data are non-identically distributed. 
To resolve this, we make use of the global statistics of the dataset. 
Because images are snapshots in time and all contain related wind physics, we allow the image-specific covariance estimation step to share information across images. This information-sharing is accomplished through regularizing covariance estimation towards the global mean. 
Simultaneously, prediction of image-specific covariance matrices is stabilized by projection into the Fourier domain, which is sparse. 

\begin{figure}
    \centering
    \includegraphics[width=\linewidth]{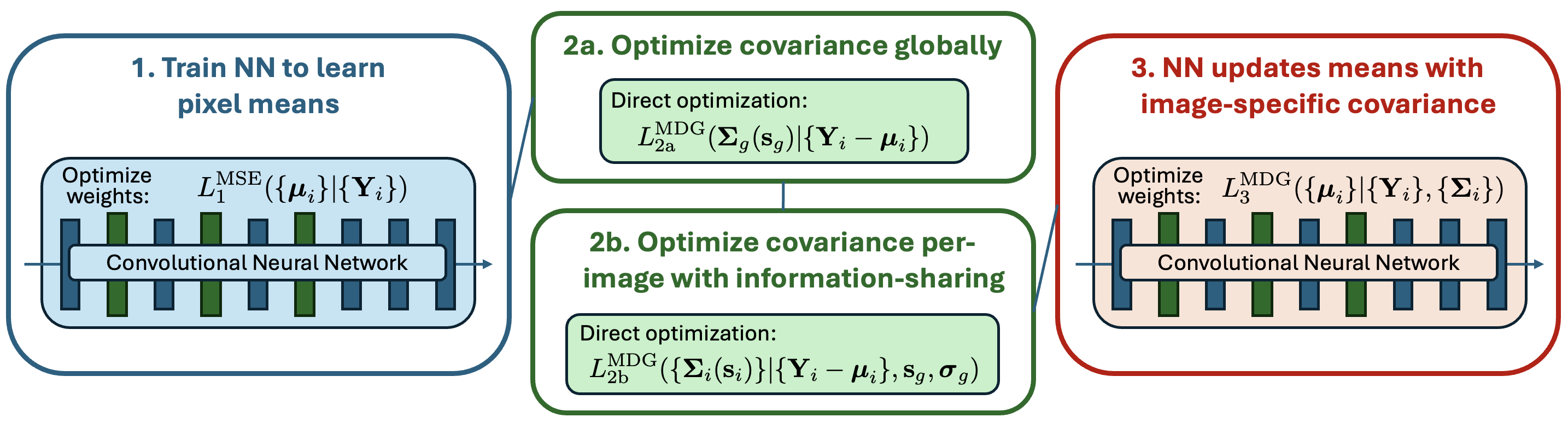}
    \caption{
        Diagrammed three-stage algorithm for multidimensional distributional neural network output for heteroscedastic data. In the loss function for each stage, variables to the left of the ``$|$" symbol are optimized over. Stage 1 involves training a CNN on a simple Mean-Squared Error loss, $L^\mathrm{MSE}_1$. Stage 2 involves optimizing the image specific covariance matrices $\boldsymbol{\Sigma}_i$ parameterized by their Fourier coefficients $\mathbf{s}_i$. To constrain the image-specific covariances, Stage 2b utilizes information-sharing regularization with hyperparameters optimized in Stage 2a. The applied regularization is equivalent to a Gaussian prior on coefficients $\mathbf{s}_i$ with mean coefficients $\mathbf{s}_g$ equal to the Fourier projected global covariance $\boldsymbol{\Sigma}_g(\mathbf{s}_g)$ (computed under the homoscedastic approximation), and image-specific prior variance $\boldsymbol{\sigma}_g^2$ defined by empirical variance of the unregularized image-specific covariance parameters from Eq.\ \eqref{eq:stage2_noreg_loss}.  
        Stage 3 reconverges the CNN from Stage 1 under the multidimensional Gaussian loss function $L^\mathrm{MDG}_3$ with image specific covariances from Stage 2 included. }
    \label{fig:algo_diagram}
\end{figure}

The algorithm is summarized as following the 3 steps illustrated in Fig.\ \ref{fig:algo_diagram}.
%% Algorithm diagram
In detail, Stage 1 optimizes the network weights ${\bf W}$ over the Mean-Squared Error (MSE) loss, 
\begin{align} \label{eq:mse_loss}
    L^\mathrm{MSE}_1(\{\boldsymbol{\mu}_i\} | \{\mathbf{Y}_i\}) = \sum_{i \in \{\mathrm{img}\}} (\mathbf{Y}_i - \boldsymbol{\mu}_i)^T  (\mathbf{Y}_i - \boldsymbol{\mu}_i),
\end{align}
in order to generate initial estimates of means $\{\boldsymbol{\mu}_i\}$ 
with $\boldsymbol{\mu}_i$ being the neural network output $f({\bf X_{i}}|{\bf W})$.

Then in Stage 2, the predicted means from Stage 1 are used to optimize 
the image-specific covariances $\{\boldsymbol{\Sigma}_i\}$ regularized around a global covariance $\boldsymbol{\Sigma}_g$. 
This process is conducted in two sub-stages: First, in Stage 2a, the global covariance, is optimized as a function of its Fourier coefficients, $\mathbf{s}_g$, by directly minimizing the multidimensional Gaussian loss function,
\begin{equation} \label{eq:stage2a_loss}
L^\mathrm{MDG}_{2\mathrm{a}}( \boldsymbol{\Sigma}_g(\mathbf{s}_g) |  \{ \mathbf{Y}_i -  \boldsymbol{\mu}_i\}) 
= 
n_s \mathrm{Tr} (\log \mathbf{s}_g) 
+
\sum_{i \in \{\mathrm{img}\}}
\mathbf{Y}_{\mathrm{err},i}^T \boldsymbol{\phi} \, \mathrm{diag}(\tfrac{n_s}{\mathbf{s}_g}) \, \boldsymbol{\phi}^\dagger \mathbf{Y}_{\mathrm{err},i},
\end{equation}
outside any network updates in a standalone maximum likelihood procedure.
The error appearing in that equation is defined by $\mathbf{Y}_{\mathrm{err},i} = \mathbf{Y}_i - \boldsymbol{\mu}_i$.
To stabilize the learning of the covariance here and in the following stages, the covariance is expanded on the discrete, 2-dimension spatial Fourier basis, $\boldsymbol{\Sigma} = \phi \, \mathrm{diag}(\mathbf{s}) \, \phi^\dagger /n_s$. This expression results from computing the sample covariance $\boldsymbol{\Sigma} \approx \mathbf{Y}^T\mathbf{Y}/n_s$ and projecting fields into the Fourier basis $\mathbf{Y}^T \approx \phi \, \mathrm{diag}({\mathbf{s^{1/2}}})$, where the last expression is approximate only because of the assumption of spatial periodicity made by the discrete Fourier transform. 
Minimization of Eq.\ \eqref{eq:stage2a_loss} has the following analytical solution, which simplifies our calculation,
\begin{equation} \label{eq:anal_sln}
    \mathbf{s}_g = \frac{\mathrm{diag}(\phi^\dagger \mathbf{Y} \mathbf{Y}^T \phi)}{n_s}.
\end{equation}

In Stage 2b, the global covariance parameters $\mathbf{s}_g$ are used to regularize the image-specific covariance parameters $\mathbf{s}_i$, effectively sharing information across the dataset, and pushing image-specific parameters towards the global values (empirically found to be the mean of the unregularized image-specific covariance parameters). The following loss is minimized directly, 
\begin{equation} \label{eq:stage2b_loss}
L^\mathrm{MDG}_{2\mathrm{b}}( \{\boldsymbol{\Sigma}_i(\mathbf{s}_i)\} | \{ \mathbf{Y}_i -  \boldsymbol{\mu}_i\}, \mathbf{s}_g, \boldsymbol{\sigma}_g) 
= 
\sum_{i \in \{\mathrm{img}\}}
\left[
\mathrm{Tr} (\log \mathbf{s}_i) 
+ 
\mathbf{Y}_{\mathrm{err},i}^T \boldsymbol{\phi} \, \mathrm{diag}(\tfrac{1}{\mathbf{s}_i}) \, \boldsymbol{\phi}^\dagger \mathbf{Y}_{\mathrm{err},i}
+ \frac{(\mathbf{s}_i - \mathbf{s}_g)^2}{\boldsymbol{\sigma}_g^2}
\right].
\end{equation}
The regularization term can be interpreted as a Gaussian prior for image-specific covariance parameters with vector standard deviation $\boldsymbol{\sigma}_g$. This standard deviation serves as hyperparameter, which we maximize in order to find the least restrictive regularization that allows the stage 3 network to generalize across images. More on this is discussed in the following sections. 

Stage 3 involves re-optimizing our CNN over the multidimensional Gaussian loss function,
\begin{equation} \label{eq:stage3_loss}
    L^\mathrm{MDG}_3(\{\boldsymbol{\mu}_i\}| \{\mathbf{Y}_i\}, \{\mathbf{s}_i\}) 
= 
\sum_{i \in \{\mathrm{img}\}}
(\mathbf{Y}_i - \boldsymbol{\mu}_i)^T \boldsymbol{\phi} \, \mathrm{diag}(\tfrac{1}{\mathbf{s}_i}) \, \boldsymbol{\phi}^\dagger (\mathbf{Y}_i - \boldsymbol{\mu}_i),
\end{equation}
which completes the first cycle of what could be considered an iterative algorithm: 1) optimizing the CNN over image means, 2) optimizing image-specific covariances with info-sharing across dataset, and finally 3) reoptimizing means for consistency with image-specific covariance matrices. Here, we stopped the analysis at a single iteration, as we observed sufficient convergence of the stage 3 network to serve as a proof-of-concept. 
%% ~~~~~~~~~~~~~~~~~~~~~~~~~~~~~~~~~~~~~~~~~~~~~~~~~~~~~~~~~~ 

%%%%%%%%%%%%%%%%%%%%%%%%%%%%%%%%%%%%%%%%%%%%%%%%%%%%%%%%%%%%%%%%%%%%%
%% Model
%%%%%%%%%%%%%%%%%%%%%%%%%%%%%%%%%%%%%%%%%%%%%%%%%%%%%%%%%%%%%%%%%%%%% 

\section{CNN Architecture and Training Specification}
\label{section:cnn}

\subsection{Architecture}

\begin{figure}
    \centering
    \includegraphics[width=\textwidth]{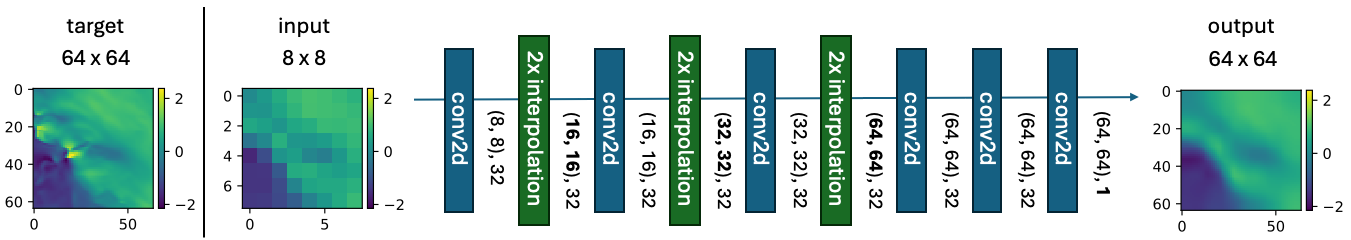}
    \caption{Diagram of the CNN architecture used throughout this work. An example set of target, low-resolution input, and high-resolution output images are shown. Image size and channel number are reported after each network layer, with bold indicating a change from output of the previous layer.}
    \label{fig:arch}
\end{figure}

State-of-the-art CNNs trained for super-resolution tasks typically consist of many layers, often utilize non-sequential data flows \cite{hu2019runet, lim2017deep}. 
Here we do not aim to outperform state-of-the-art super-resolution performances, but rather we demonstrate a proof-of-concept for probabilistic outputs of feed-forward networks tractable enough to ensure ease of analysis and training consistency. After analysis of various network depths and convolutional layer feature numbers, we settle on the following minimal network capable of training on both multidimensional Gaussian loss and MSE loss to better super-resolution performance than interpolation. The base network consists of 6 convolutional layers, each with 32 channels (features). The up-scaling is done by a nearest-neighbor interpolation before the 2nd, 3rd, and 4th convolutional layers. 
We tried different configurations of the interpolation and convolutional layers, and found that staged interpolation early in the network performed best. 
This architecture is diagrammed in Fig.\ \ref{fig:arch}.

\subsection{Data}\label{sec:data}
Surface wind speed is extracted from high-resolution, convection-permitting model simulations produced as part of the UK Natural Environment Research Council (NERC) Cascade project \cite{pearson2010,love2011,holloway2012}.  
We use the 4~km resolution Cascade simulation over the tropical Indo-Pacific Warm Pool. 
A detailed description of the simulation is presented in \citeA{holloway2012}.   
The simulation begins on 6 April 2009 and spans 10 days, chosen as a case study of an active Madden-Julian oscillation (MJO) event. 
The data are stored at full resolution in space and once an hour in time ($214$ time snapshots). 
Thorough validation of the Cascade simulation has been reported \cite{holloway2012,holloway2013,holloway2015}. 
Surface wind speed was taken from a set of 4 neighboring (non-overlapping) spatial subregions of 64 x 64 grid-points in the central Southern Indian Ocean between longitudes $77.9^{\circ}E$ and $82.5^{\circ}E$ and latitude from $-6.4^{\circ}S$ to $-11^{\circ}S$. 
This results into 856 ($=4\times214$) 64 $\times$ 64 grid-point high-resolution snapshots of surface wind speed. 
We subsample these images to 8 $\times$ 8 grid-point low-resolution inputs.
Finally, the data were normalized to zero mean and unit standard deviation across the combined training and testing sets.

\subsection{Training}\label{sec:training}
The CNNs in Stage 1 and 3 are trained to 300 epochs, where we found losses to consistently plateau. The stage 3 network starts with weights converged upon in Stage 1. Convolutional layers are padded with edge grid-point replicated across the kernel span. Network construction and training was implemented in \texttt{PyTorch} using the Adam optimizer. For the stage 1 MSE optimization, the learning rate was fixed at $10^{-2}$. For the stage 3 MDG optimization, the learning rate was set to decay from $10^{-2}$ to $10^{-4}$ at a decay rate of 0.95.  The total dataset was 856 images split by 75\% train and 25\% test, resulting in 642 training images and 214 test images. The dataset used was a timeseries of 214 hours for the entire region. To build the train/test split, the data were ordered by subregion and then in time, with the training data being the first 3/4 of the timeseries for each subregion and the test set being the last 1/4 of the timeseries for each subregion.

%%%%%%%%%%%%%%%%%%%%%%%%%%%%%%%%%%%%%%%%%%%%%%%%%%%%%%%%%%%%%%%%%%%%%
%% Model output
%%%%%%%%%%%%%%%%%%%%%%%%%%%%%%%%%%%%%%%%%%%%%%%%%%%%%%%%%%%%%%%%%%%%% 
\section{Stage-Wise Results and Analysis}
\label{section:results}

In this section, we analyze the output from each of the 3 stages within our proposed framework. Afterwards, we generate image-specific samples using the output from the full framework and compare statistical properties of the Image-Specific Covariance model to the Global Covariance model (where Stage 2b of the framework is skipped). Before proceeding with results, we first define and discuss our chosen performance metric.

Much literature exists on benchmarking super-resolution techniques and the relevance of common metrics for assessing super-resolution model performance \cite{yang2014single}. But most of the literature concerns super-resolution on images of natural scenes, and metric efficacy is assessed with the axiom of human perception being the ultimate judge of image reconstruction accuracy. Here, our task is different. We are interested in super-resolving outputs from earth system models in order to aid further computation and potentially visualization. 
In the case of visualizing super-resolved atmospheric variables (such as surface wind speed), then perceptual accuracy of the super-resolved fields would be a priority. 
But in the case of using the super-resolved fields in subsequent analysis and computation, we have no reason to bias metrics around human visual perception. 

To keep our performance metric as simple and interpretable as possible, we have chosen the Mean Absolute Percent Error (MAPE) defined by the mean point-wise absolute accuracy,
\begin{equation}
    \mathrm{MAPE}(\mathbf{Y}, \boldsymbol{\mu}) = \sum_{i=1}^{n_p} \left| \frac{Y_i - \mu_i}{Y_i}\right|\frac{1}{n_p}.
\end{equation}
We choose this metric for its ease of interpretation and general quantification of image replication at the grid-point--by--grid-point level, and accept that the MAPE may overweight small changes in the image mean and underweight perceptually distinct features at small length scales. 
In what follows, we compare the model performance with the MAPE computed on output images and on image gradients. We assess probabilistic model outputs with more comprehensive surface boxplots.

\subsection{Stage 1: Train CNN on MSE loss}

\begin{figure}
\centering
\noindent\includegraphics[width=.65\linewidth,angle=0]{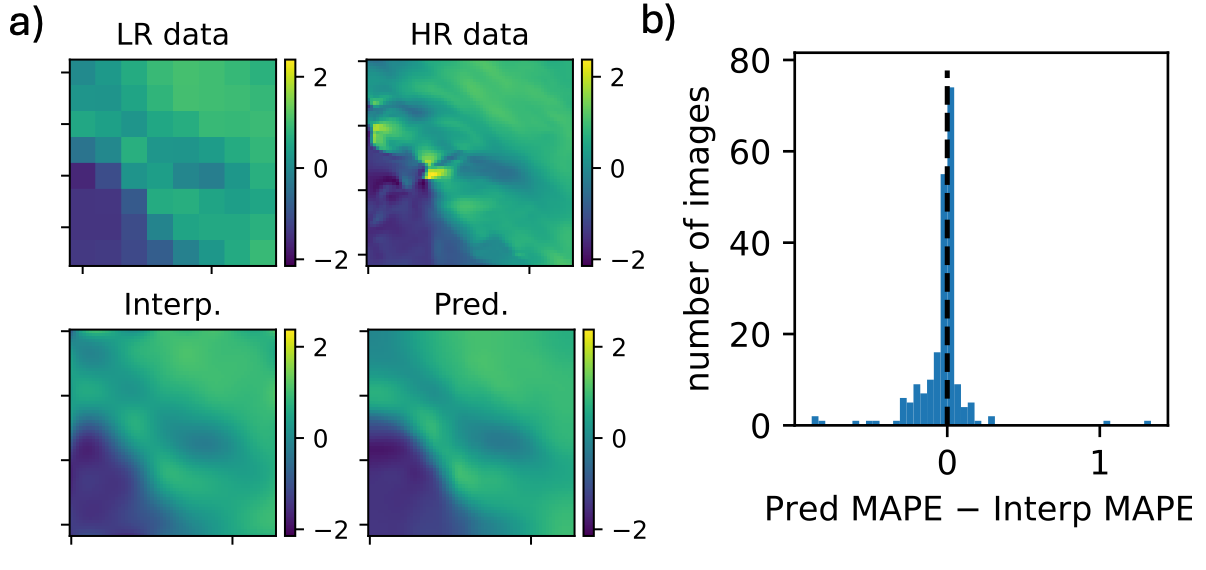} 
\caption{
    Summary of stage-1 output: MSE trained model performance.
    Panel (a) shows a single test image example, comparing the low resolution input (LR), the predicted high resolution output (Pred.), a cubic interpolation (Interp.) of the LR input, and the HR target image. Panel (b) visualizes the image-wise comparison of prediction and interpolation MAPE in a histogram. Each image is binned by the prediction error minus interpolation error, so values less than zero are predictions better than interpolation. 
    }\label{fig:mse_performance}
\end{figure}

Stage 1 of the model results from training the CNN on a standard MSE loss. For all test images, the MSE model performs better than cubic interpolation. Panel (a) of Fig.\ \ref{fig:mse_performance} shows a single test image example, comparing the predicted high-resolution output to a cubic interpolation of the LR input as well as the HR target image. 
It is clear that much detail is lost in the CNN prediction, but the network does recover more details than the interpolation. 
In assessing performance, we must note the downscaling factor is 8, which is large in the realm of recent downscaling applications \cite{wang2020deeplearningimagesuperresolution, tian2020coarse,merizzi2024wind}. 
Panel (b) of Fig.\ \ref{fig:mse_performance} aggregated image-wise performance across the test set. The histogram compares the prediction and interpolation MAPE for each image, which is centered near zero and comparable to interpolation due to the high downscaling factor of 8.   

\subsection{Stage 2: Direct Optimization of MDG Loss over Image-Specific Covariance Fourier Parameters}

\begin{figure}
\centering
\noindent\includegraphics[width=.65\linewidth,angle=0]{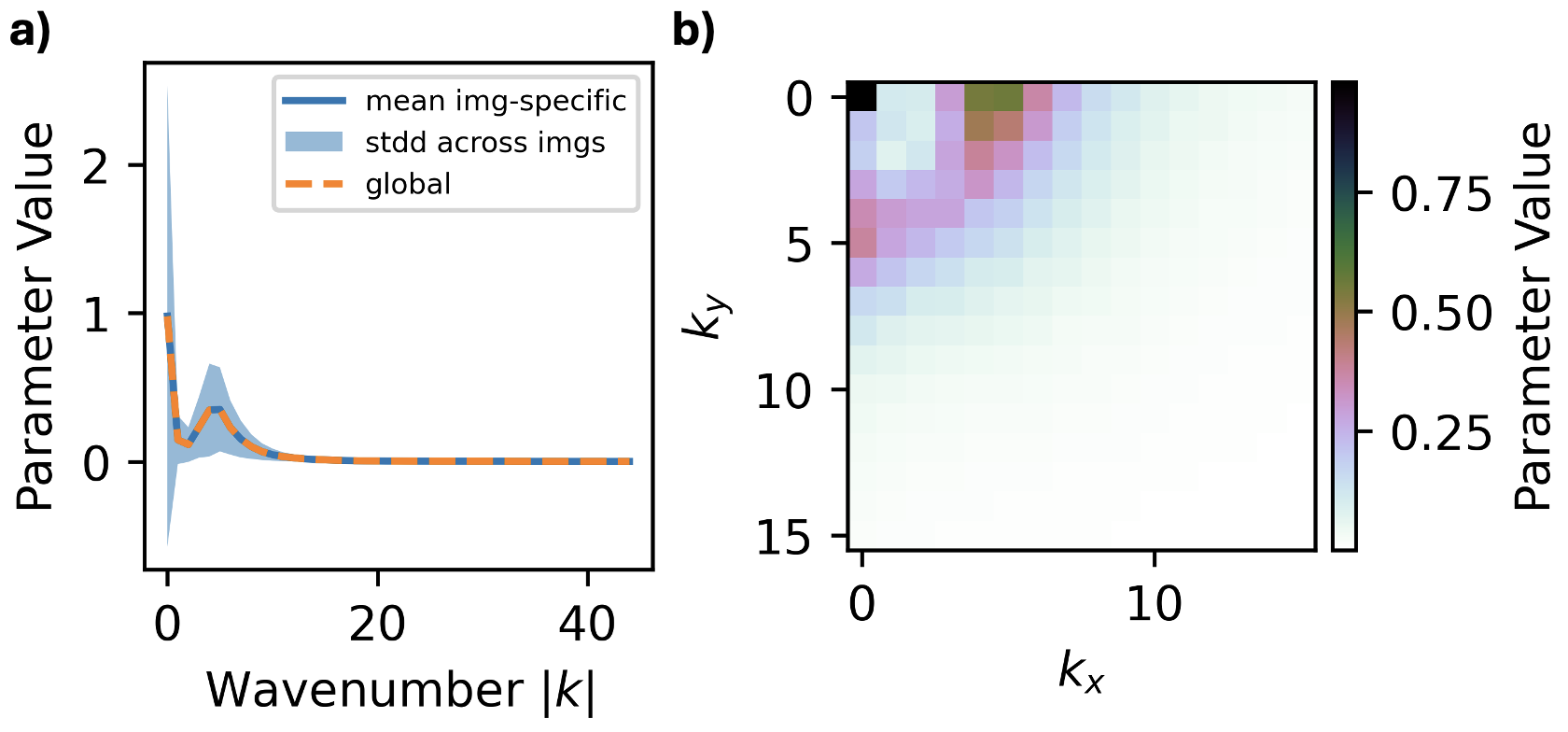}
\caption{
    \label{fig:params_by_k}
    Global and unregularized image-specific covariance parameters as functions of Fourier space wavevector $\mathbf{k}$. 
    In Panel (a) (left), the image-specific covariance parameters are compared to global Fourier parameters $\mathbf{s}_g$ as a function of their wavenumber $|k|$. Intuitively, the mean across images $\langle\mathbf{s}_i^\mathrm{unreg.}\rangle_i$ is equal to $\mathbf{s}_g$. The standard deviation of covariance parameters across images is plotted in blue fill, demonstrating high dispersion image--to--image. In Panel (b) (right), the image-specific parameters are plotted as a function of the wavevector $\mathbf{k} = (k_x, k_y)$. The peak seen in Panel (a) at $|k| \approx 5$ is seen in Panel (b) to have approximate circular symmetry. 
    }
\end{figure}

Stage 2 takes as input the predicted means output from Stage 1 and outputs image-specific covariance matrices 
Although it is ill-conditioned to estimate distributional parameters from single samples, regularizing the image-specific covariance matrices in Fourier space makes it possible.  
Before implementing the information-sharing regularization term in Eq.\ \eqref{eq:stage2b_loss}, we directly minimized the loss
\begin{equation} \label{eq:stage2_noreg_loss}
L^\mathrm{MDG}_{2\mathrm{b}\text{, no reg.}}( \{\mathbf{s}_i\} | \{\mathbf{Y}_i -  \boldsymbol{\mu}_i\}) 
= 
\sum_{i \in \{\mathrm{img}\}}
\left[
\mathrm{Tr} (\log \mathbf{s}_i) 
+ 
\mathbf{Y}_{\mathrm{err},i}^T \boldsymbol{\phi} \, \mathrm{diag}(\tfrac{1}{\mathbf{s}_i}) \, \boldsymbol{\phi}^\dagger \mathbf{Y}_{\mathrm{err},i}
\right],
\end{equation}
which has the analytic solution per-image following Eq.\ \eqref{eq:anal_sln}.
These unregularized image-specific covariance parameters $\{\mathbf{s}_i^\mathrm{unreg.}\}$ are plotted in Fourier space in Fig.\ \ref{fig:params_by_k}. In Panel (a), the image-specific covariance parameters are compared to global parameters $\mathbf{s}_g$ as a function of their wavenumber $|k|$. Intuitively, the mean across images $\langle\mathbf{s}_i^\mathrm{unreg.}\rangle_i$ is equal to $\mathbf{s}_g$. 
The image-wise standard deviation of covariance parameters is plotted in blue fill within Fig.\ \ref{fig:params_by_k}a, showing high dispersion image--to--image. 
In Panel (b), the image-specific parameters are plotted as a function of the wavevector $\mathbf{k} = (k_x, k_y)$. The peak seen in Panel (a) at $|k| \approx 5$ is seen in Panel (b) to have approximate circular symmetry. 
Interestingly, for images across the train and test sets, Fourier modes of wavenumber 0 and $5 \pm 2 $ contribute most of the covariance. 
Possibly, there is a potential to make more use of this structure in training a more flexible algorithm, such as in low-rank approximations that may stabilize the network in Stage 3 or a parametric shape to model frequency contribution as seen in Panel (b). 
For now, we leave this finding for future work. 

Notably, Stage 2 in the form implemented here requires ground-truth image data due the dependence of the loss on $\mathbf{Y}_\mathrm{err}$. Although this makes the entire 3-stage algorithm unable to predict out-of-sample, we aim here is to prove the concept of CNNs with distributional output and leave to future work development an out-of-sample prediction method for $\boldsymbol{\Sigma}$.  Likely, this prediction could be performed accurately by training a simple neural network on the $\boldsymbol{\Sigma}$ values  derived here. Note that the proposed Global-covariance framework can handle out-of-sample prediction as is.

\begin{figure}
\centering
\noindent\includegraphics[width=.65\linewidth,angle=0]{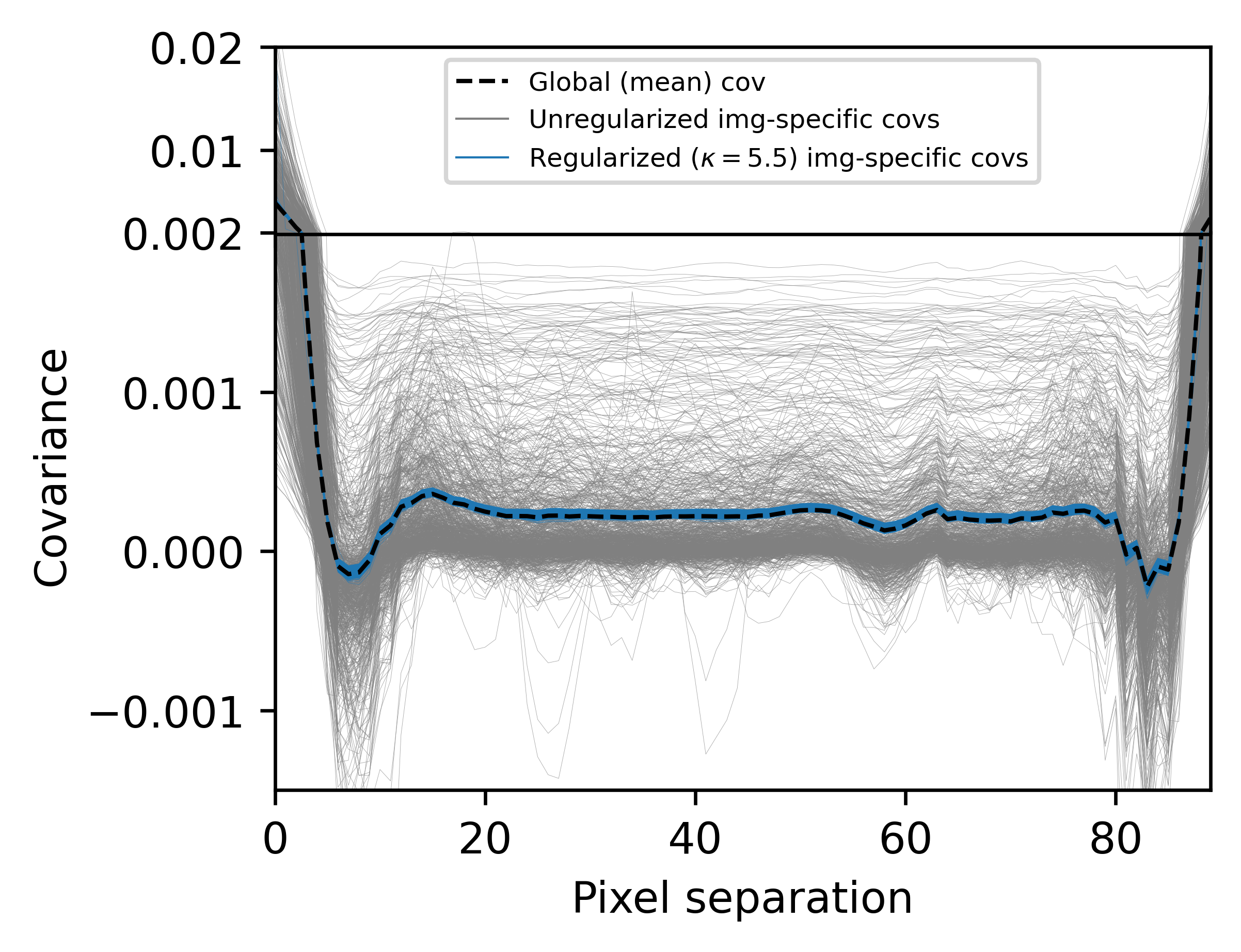}
\caption{
    \label{fig:img_spec_covs}
    Covariance estimates shown as function of grid separation. 
    Unregularized image-specific covariances are plotted in gray.
    The globally optimal covariance (equal to the mean of image-specific covariances) is plotted in dashed black. 
    Image-specific covariances computed with optimal scale-factor $\kappa = 5.5$ are plotted in blue and appear solid, but when compared to the dashed black reveal a small spread. 
    Each of the covariance functions plotted here was computed by computing the grid-point separation distance corresponding to each element of the covariance matrix and averaging values at identical grid separations (rounded to the nearest integer number of grid-points).  
    The solid black horizontal bar marks a split between y-axis scales, to better visualize both diagonal values of the covariances and variation at middle distance grid-point separations.
    }
\end{figure}

Covariance matrices per image were generated by inverse Fourier transform of predicted parameters. 
In Fig.\ \ref{fig:img_spec_covs}, various covariance matrices are visualized by computing covariance as a function of grid-point separation distance.
Unregularized image-specific covariances are plotted in gray, and show similar qualitative form despite high variation in magnitude of diagonal values (grid separation 0) and uncorrelated plateau (flat section in at middle distance separations beyond spatial the correlation length). 
The globally optimal covariance (equal to the mean of image-specific covariances) is plotted in dashed black. 
Image-specific covariances computed with optimal scale-factor $\kappa = 5.5$ are plotted in blue. 
The condensed blue lines appear solid, but when compared to the dashed black reveal a small spread. This reveals the significant reduction in covariance dispersion needed to constrain the stage 3 network.  
Values at 0 grid separation indicate diagonal values of the covariance, and fall with expected spatial correlation length of approximately 7 grid-points. 
The peak in values at large separations is an anomaly of the Fourier basis, which enforces spatial periodicity. The number of grid-points at high separations is low (only covering corner-to-corner correlation), so we choose to accept this anomaly of the Fourier basis at the benefit of sparse representation and training stability. 

As will be discussed in the following section on results from Stage 3, the unregularized covariance parameters lead to a Stage 3 CNN that collapses to predicting uniform images with values near the target image mean. However, using the global covariance that minimizes Eq.\ \eqref{eq:stage2a_loss} directly in Stage 3 leads to stable predictions. 
In order to stabilize the training in Stage 3 with image-specific covariance matrices, we found success in regularizing the parameters as shown in Eq.\ \eqref{eq:stage2b_loss}. The regularization choice is equivalent to placing a Gaussian prior on the Fourier parameters $\{ \mathbf{s}_i \}$ with mean equal to the global optimum $\mathbf{s}_g$. The variance of the Gaussian prior was set to the variance of the unregularized image-specific parameters, modified by a dispersion reducing scale factor, $\boldsymbol{\sigma}_g = \sqrt{\mathrm{Var}(\mathbf{s}_i^\mathrm{unreg.})} / \kappa$. The scale factor $\kappa$ is treated as a hyperparameter and optimized by finding the minimum value that allowed the stage 3 network to converge to satisfactory predictions on the test set. To perform the hyperparameter sweep over $\kappa$, Stages 2b and 3 were trained with values ranging from 10 through 1 at intervals of 0.5. The smallest $\kappa$ value that allowed Stage 3 to converge was considered optimal; this was found to be $\kappa = 5.5$.

\subsection{Stage 3: Retrain CNN on MDG Loss with Image-Specific Covariances}

In Stage 3, we continue training the CNN from Stage 1 with a MDG loss including information contained in the image-specific covariances estimated in Stage 2. 
As discussed above, using the covariance parameters that result from optimization of the stage 2 loss without any regularization caused the stage 3 network output to collapse onto the mean value of each image. 
We interpret this result as a failure of the CNN to generalize any information between independently distributed images. 
The difference between image distributions can be visualized in part by noting the image-specific covariances in Fig.\ \ref{fig:img_spec_covs}. 
It is especially important to note the difference in diagonal and off-diagonal values, %of different covariance, 
indicating differing levels of local variance and long-range spatial correlation.  

\begin{figure}
    \centering
    \includegraphics[width=1.0\linewidth]{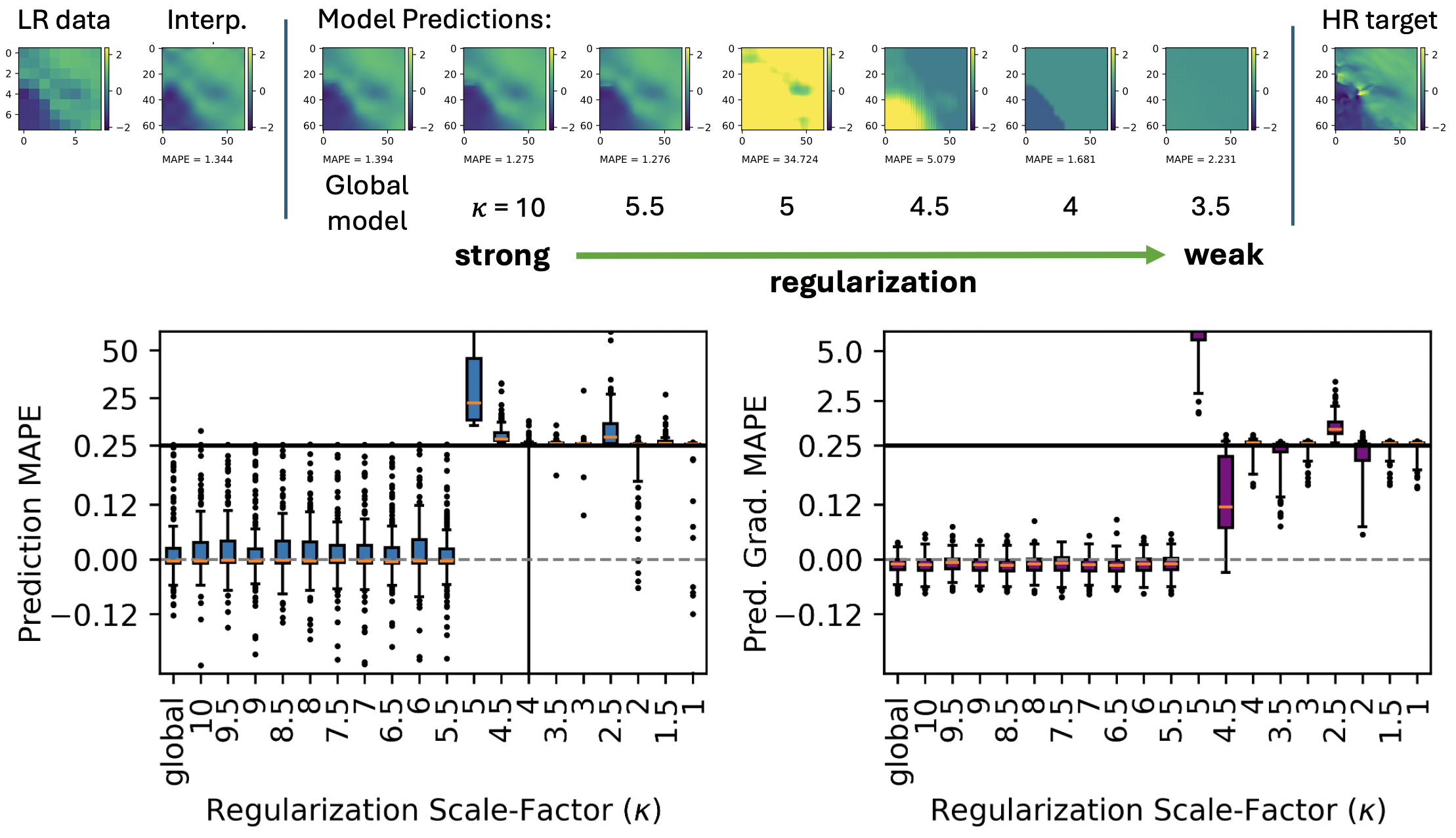}
    \caption{
        \label{fig:kappa_comp}
        Model performance across values for hyperparameter $\kappa$ controlling dispersion of image-specific covariances. 
        Decreasing values of $\kappa$ therefore indicate decreasing regularization strength on Fourier coefficients of the image-specific covariances.
        Images at the top display a single test image (from left to right): low-resolution input data, interpolation outputs, Stage 3 predictions across models with varying $\kappa$, and the high-resolution target image. 
        The boxplots display model performance over the entire test set for the Global Covariance model and Image-Specific covariance model with different values of $\kappa$. Boxplots on the left consist of the MAPE computed on each test image relative to stage 1 output (MSE trained CNN). Therefore, the zero point on the y-axis is a stage-3 image prediction with error equal to its MSE equivalent. 
        Right boxplots include the same analysis but with MAPE computed on image gradients and plotted relative to image gradient MAPE for MSE model predictions. 
        }
\end{figure}

To tune the image-specific covariance scale-factor $\kappa$, we optimized the stage 3 network on outputs from Stage 2b with different $\kappa$ values. 
Fig.\ \ref{fig:kappa_comp} illustrates this hyperparameter sweep, with a single test image prediction corresponding to different dispersion scale factors. 
It is clear that between $\kappa = 5.5$ and $\kappa = 5$, the network predictions collapse to predicting a uniform image with value near the target mean. 
We conjecture that the network fails to generalize spatial details at $\kappa = 5$ because the image-specific distributions have become too dissimilar. 
Past this threshold, the network is not able to learn any spatial details. 
These results suggest a phase change in network with decreasing similarity between image distributions, a phenomenon that is left  subject of future work. 

The lower panel of Fig.\ \ref{fig:kappa_comp} compares model performance over the entire test set for the Global Covariance model and Image-Specific covariance model with different values of $\kappa$. Each boxplot on the left consists of the MAPE computed on each test image relative to Stage 1 outputs (MSE-trained CNN). Therefore, the zero point on the y-axis is a stage-3 image prediction with error equal to its MSE equivalent. Negative values indicate better performance than MSE.

The right panel of purple boxplots in Fig.\ \ref{fig:kappa_comp} visualizes computed MAPE over image gradients for evaluation of prediction accuracy in regions of high variability. 
Models with $\kappa \geq 5.5$ perform slightly better than their MSE-trained equivalent (median MAPE is less than zero). 
Interestingly, this set of models seems to perform even better in terms of image gradient. 
Based on the spike in MAPE values at $\kappa = 5$ for both the predictions and prediction gradients, we identify 
$\kappa = 5.5$ as the least amount of regularization required to converge Stage 3 to realistic predictions. 
The image-specific covariances for the $\kappa = 5.5$ model are plotted in blue in Fig.\ \ref{fig:img_spec_covs}.

\subsection{Analysis of Statistical Output}

\begin{figure}
    \centering
    \includegraphics[width=0.8\linewidth]{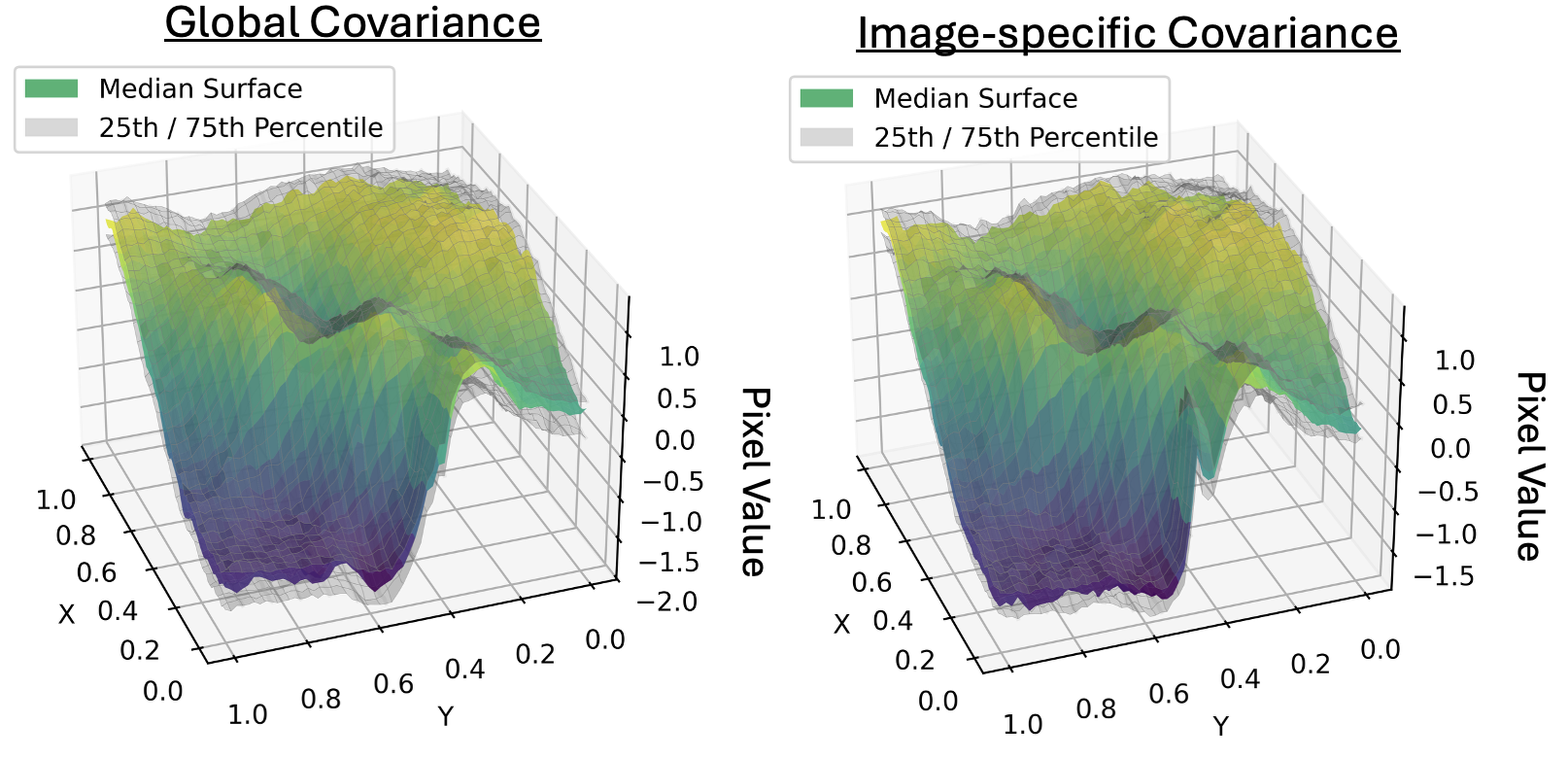}
    \caption{
        \label{fig:surface_boxplots}
        Surface boxplots computed over samples generated for a single test image (out-of-training sample) for the Global Covariances (top) with $\kappa = 5.5$ Image-Specific Covariance model (bottom). For each model 100 samples were generated from the MDG distribution used as a loss in Stage 3 with appropriate covariances. 
        Then, each sample is sorted by centrality using the band depth measure discussed in the main text. The central 50\% region of the sample surfaces are shown in transparent gray. The median surface is colored with the same blue-green-yellow color map used throughout this manuscript. 
        }
\end{figure}

We use the Gaussian likelihood \eqref{eq:mvn} to generate statistical samples and provide samples reflecting the uncertainty about our predicted ``mean'' images. 
In particular, we generated 100 samples for each predicted image under the Global Covariance and Image-Specific Covariance models and generated surface boxplots over these samples. Surface boxplots are two-dimensional functional boxplots that are a generalization of standard boxplots used broadly in data visualization. For functional data, quantifying the spread in data around a median value requires computing the metric of functional centrality, referred to as band depth, see \cite{lopez2009concept}. We utilize the \texttt{scikit-fda} \texttt{Python} package \cite{ramos-carreno++_2024_scikit-fda,ramos-carreno++_2024_scikit-fda-repo}) to compute the sample depth and generate surface boxplots. 
Surface boxplots for a single test image are shown in Fig.\ \ref{fig:surface_boxplots} for both the Global Covariance model and the  Image-Specific Covariance model with $\kappa = 5.5$. The central 50\% region of the sample surfaces are shown in transparent gray. The median surface is visualized with the same blue-green-yellow color map used throughout the manuscript, see Fig.\ \ref{fig:kappa_comp}. 
Comparing surfaces between the two models, it is clear that the difference between results is minimal. This is consistent with the high degree of regularization on image-specific covariances needed to converge Stage 3 (see blue $\kappa = 5.5$ model covariances in Fig.\ \ref{fig:img_spec_covs}). 

\begin{figure}
    \centering
    \includegraphics[width=0.65\linewidth]{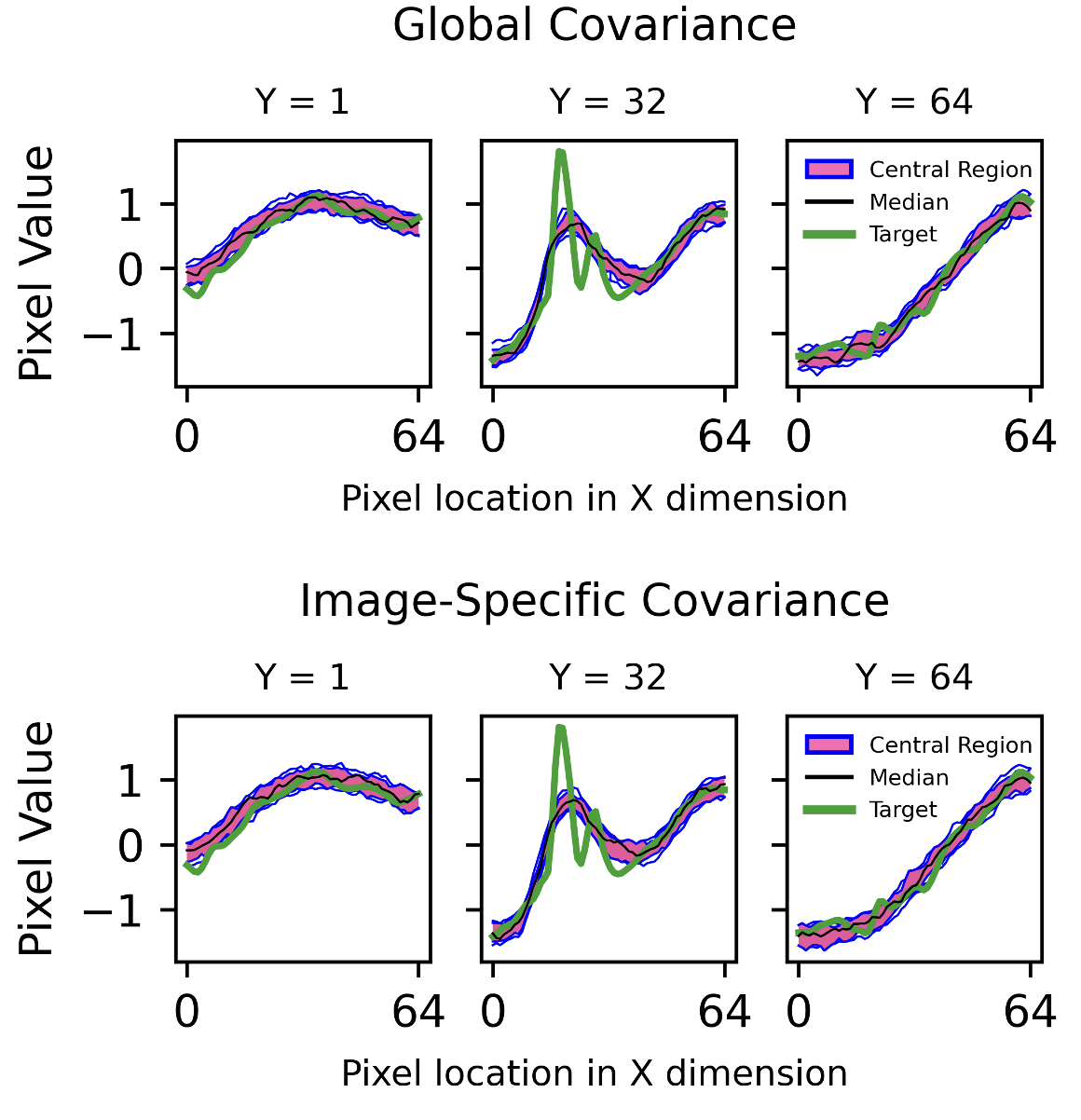}
    \caption{
    \label{fig:boxplot_slices}
    Constant Y slices of the surface boxplots in Fig.\ \ref{fig:surface_boxplots}. The central region is filled with pink and outlined in blue. Outlier boundaries defined by the standard 1.5 multiple of the central region depth are also plotted in blue, but lie very close to the edge of the central region. The median sample is plotted in black. Constant Y slices of the target image are plotted in green. 
    The corresponding quantitative analysis of this coverage of the target image with the samples' central region is included in Fig.\ \ref{fig:coverage}.
    }
\end{figure}

To better visualize quantitative differences between the model statistical outputs, Fig.\ \ref{fig:boxplot_slices} visualizes 1-dimensional slices through the 2D surface boxplots. The 3 surface boxplots slices are taken at each edge of the image (Y grid-point 1 and 64) and through the center of the image (Y grid-point 32). 
In Fig.\ \ref{fig:boxplot_slices}, the median sample is plotted in black and the central 50\% region of the sampled surfaces is plotted in pink and outlined in blue. 
Also in blue, the standard outlier boundary is plotted (1.5 times the central region depth). Corresponding slices of the target image are plotted in green. 
For both models, we see that regions of the target test image with lower variability (the two edges in this case, plotted on left and right) are covered well by the central region of the functional boxplots. 
The central Y slice of the target image (plotted center) shows higher variability and strays significantly from the boxplot central region. 
From this we conclude that both models are under-dispersed estimates of uncertainty. This is consistent with the high degree of regularization needed on image-specific covariances: The image-specific model is actually quite similar to the global covariance model. Future work will address this limitation. 
Figure \ref{fig:boxplot_slices} facilitates the validation of the predicted distribution with a single sample. The expectation for an optimal model tested on a single sample would be that 50\% of the target image slide would lie inside the central region of the functional boxplot. 
Here, we see similar results with the Global Covariance model and the Image-Specific Covariance model, with high ``coverage'' of the target image slices with the central regions at image edges (left and right) and slightly less coverage in the more highly variable middle slice. 

\begin{figure}
    \centering
    \includegraphics[width=0.65\linewidth]{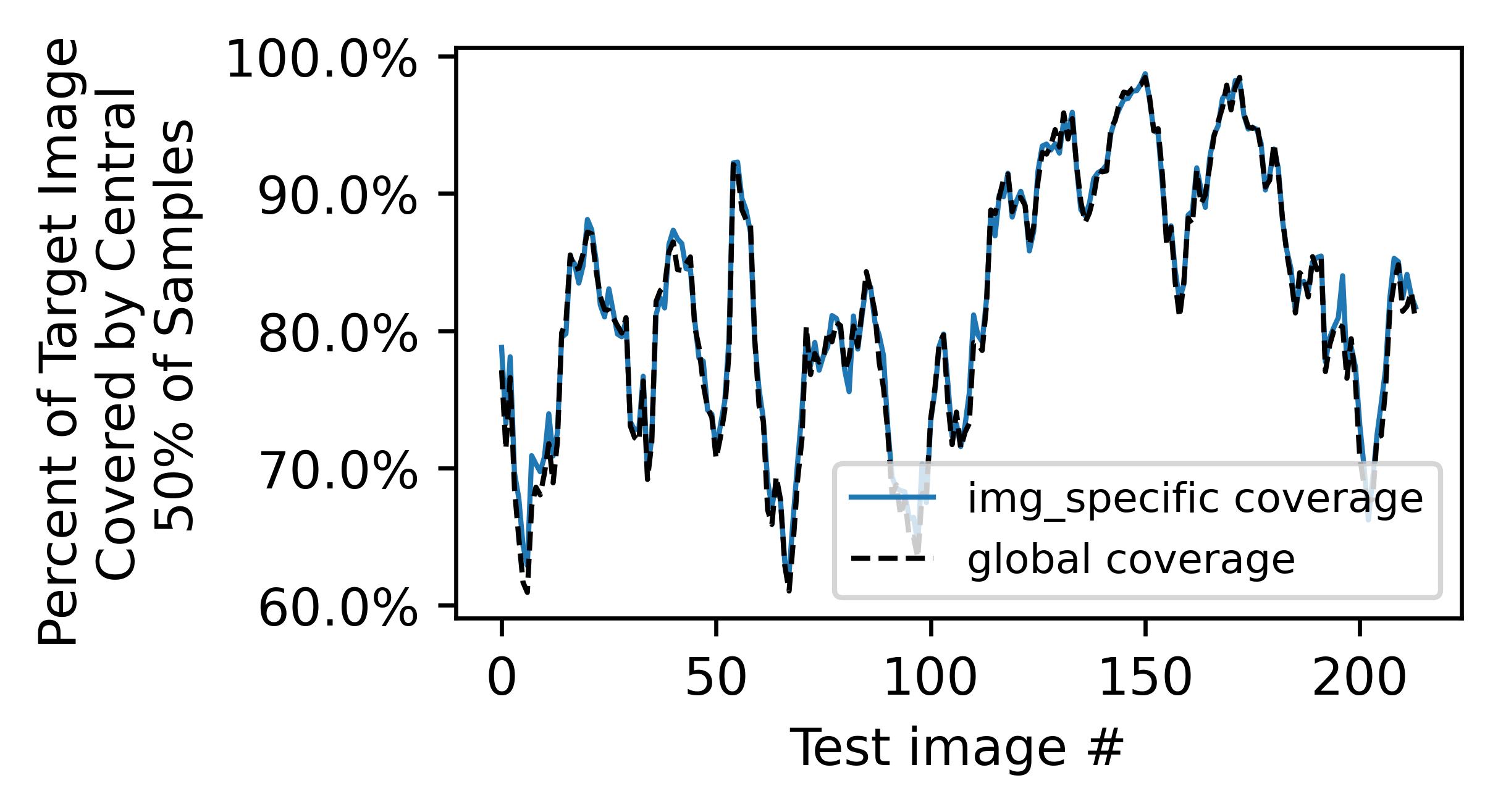}
    \caption{ \label{fig:coverage}
        Coverage percentages of target image by central region of surface boxplot for each image in the test set, computed for both the Global Covariance (dashed black) and $\kappa=5.5$ Image-Specific Covariance (blue) models. If the models optimally captured predicted image distributions, the expectation would be that 50\% of the target image slide lie inside the central region of the functional boxplot.  The ordering of the test data has been changed for visualization clarity. The test data is plotted as sequential subregion-specific timeseries.
        }
\end{figure}

To perform a quantitative analysis of the statistical outputs of the model in the test data, we computed the coverage of the target image by the 50\% central region of the surface boxplot for each image in the test set. 
Fig.\ \ref{fig:coverage} plots coverage across the test set for both the Global Covariance (dashed black) and $\kappa=5.5$ Image-Specific Covariance (blue) models. 
Coverages for both models track each other, consistent with their similar performance seen in Fig.\ \ref{fig:kappa_comp}. 
Since this is effectively a validation of distributions for each image on a single sample, we would expect 50\% coverage. 
The curves oscillate from 60\% to nearly 100\%, demonstrating that predictions for every test image are underdispersed to varying degrees. 
As discussed earlier, this apparent underdispersion of the models could result from several factors, such as the high degree of constraints in the regularization of image-specific covariance parameters that we found necessary for Stage 3 to converge.  
Relaxing the regularization would require other strategies to allow the Stage 3 network training to stabilize, including potentially reworking the architecture with skip connections or enforcing physics-based constraints for the given application.  

\section{Conclusion}
\label{section:conclusion}

In this work, we present a framework for learning closed-form, multidimensional probabilistic output from neural networks trained on non-identically distributed, heteroscedastic data. 
Our iterative technique is flexible and broadly applicable across domains. 
For this proof--of--concept with a multidimensional Gaussian loss, we showed that learning image-specific covariance matrices is feasible when leveraging the numerical stability of the naturally sparse Fourier representation of smooth spatial data. 
While directly training on these per-image covariances introduces challenges, we addressed this with a regularization scheme that pulls image-specific covariance estimates toward a global, homoscedastic approximation: a strategy we refer to as information sharing. 
Our results suggest the potential for closed-form, uncertainty-aware modeling of complex spatial data, and we suggest this framework as a viable approach to generating multidimensional distributional output from feed-forward neural networks without loss of mean prediction performance. 
This framework opens the door to efficient sampling of interpretable neural network output with possible extension to more expressive distribution families, including tail-embedding models. Specifically, the proposed framework of probabilistic losses can be applied to other distributions provided that they have an explicit location (mean) parameter. However, there exist very few tractable multidimensional probability distributions. Future work could explore the use of copula models combined with flexible bulk-and-tails marginal distributions \cite{stein2021parametric} to model distributions beyond Gaussian assumptions.

\section{Future Work} \label{section:future}

This work represents an early-stage proof-of-concept, and several steps remain before the framework can be utilized in contexts with high need for accurate uncertainty specification. 
While we have demonstrated the feasibility of training neural networks to output closed-form multidimensional distributions without sacrificing performance offered by more tried-and-true deterministic modeling methods. We note that our performance evaluation is based on a single sample per prediction making it difficult to draw strong conclusions about the fidelity of the learned distributions. 
A more robust evaluation strategy will be needed to assess predictive calibration and sharpness.

A key component of future work lies in refining the information-sharing based regularization scheme introduced in Stage 2. 
While our current approach of pulling image-specific covariances toward a global homoscedastic baseline helps stabilize training, it likely suppresses meaningful dispersion in the data. 
A more nuanced strategy is needed to better balance accurate uncertainty quantification with convergence stability. 
Relatively simple solutions could be constructed by regularizing to means specific to spatial regions, or regularizing to means across several images neighbored in time. 
More complex regularization strategies for Stage 2 could involve different distributional forms for the prior on Fourier parameters of the covariance.  
It is also worth exploring whether Stage 3 could be regularized in ways that avoid the need for strong constraints on Stage 2, while still enabling the network to learn valid, high-dimensional covariance structures.

Another limitation is that Stage 2 is currently designed strictly for in-sample reconstruction. 
Extending this to support out-of-sample prediction would be relatively straightforward from an architectural perspective but opens a separate set of research questions around generalization, especially under non-identically distributed statistics \cite{neyshabur2017exploringgeneralizationdeeplearning}.

The tendency of Stage 3 to collapse into predicting only the image means at small values of the regularization parameter $\kappa$ remains poorly understood. 
This phase transition behavior raises theoretical questions about the geometry of the loss landscape and will be investigated in future work. 
It is possible that changes to the network architecture would facilitate better generalization across heteroscedastic data.

Another question left aside from this work is epistemic uncertainty. Resolving this missing component of uncertainty quantification could be achieved with a comprehensive review of model performance across different samples or with some form of Bayesian model averaging across an ensemble of predictions. All of this work could also be extended to Bayesian neural networks. 

Together, these directions point to a rich space of open problems, which goal is to build on this foundation to develop closed-form distributional neural networks expressive enough to capture uncertainty around NN outputs with complex distributions.

\section*{Open Research Section}
Surface wind speed data used for this work can be obtained from the Cascade project is available on request from the NERC Centre for Environmental Data Analysis (CEDA) \cite{nerc2008cascade}.
The python software and data used to generate results in this paper is available at \url{https://github.com/HarrisonGoldwyn/distributional_SRCNN}.

\acknowledgments
This work was authored by the National Renewable Energy Laboratory for the U.S. Department of Energy (DOE) under Contract No. DE-AC36-08GO28308. Funding provided by Department of Energy Office of Science Advanced Scientific Computing Research, DOE Award DE-SC0024721. The research was performed using computational resources sponsored by the DOE Office of Energy Efficiency and Renewable Energy and located at the National Renewable Energy Laboratory. The views expressed in the article do not necessarily represent the views of the DOE or the U.S. Government. The U.S. Government retains and the publisher, by accepting the article for publication, acknowledges that the U.S. Government retains a nonexclusive, paid-up, irrevocable, worldwide license to publish or reproduce the published form of this work, or allow others to do so, for U.S. Government purposes.

GPT-4o was used to generate editing suggestions to nontechnical portions of this manuscript for improved readability of prose. No text generated by GPT-4o was used in the writing of this manuscript without careful review and edits by the authors. All use of GPT-4o in preparation of this publication was limited to use consistent with NREL publication policy and AGU policy on generative AI.

%%%%%%%%%%%%%%%%%%%%%%%%%%%%%%%%%%%%%%%%%%%%%%%%%%%%%%%%%%%%%%%%%%%%%
% REFERENCES
%%%%%%%%%%%%%%%%%%%%%%%%%%%%%%%%%%%%%%%%%%%%%%%%%%%%%%%%%%%%%%%%%%%%%
\bibliography{srcnn}

% ~~~~~~~~~~~~
\end{document}